\documentclass[10pt,twocolumn,letterpaper]{article}

\usepackage{wacv}
\usepackage{times}
\usepackage{epsfig}
\usepackage{graphicx}
\usepackage{amsmath}
\usepackage{amssymb}
\DeclareUnicodeCharacter{00A0}{ }
\DeclareUnicodeCharacter{00A0}{~}
\usepackage[utf8x]{inputenc}

\usepackage{multirow}
\usepackage{wrapfig}
\usepackage[colorlinks]{hyperref}
\usepackage[linesnumbered]{algorithm2e}
\usepackage{algorithmic}
\usepackage{caption}
\usepackage{subfigure}
\usepackage[font=small,labelfont=bf]{caption}

\usepackage{epstopdf}
\usepackage{graphicx}
\usepackage{subfigure}
\usepackage{enumitem}

\wacvfinalcopy 

\begin{document}

\title{Deep Image Compositing}
\author{He Zhang$^{*1}$, Jianming Zhang$^{*1}$, Federico Perazzi$^{*1}$, Zhe Lin$^{*1}$, Vishal M. Patel$^{*2}$ \\
1. Adobe Research  \quad
2. Johns Hopkins University \\
}

\makeatletter
\let\@oldmaketitle\@maketitle
\renewcommand{\@maketitle}{\@oldmaketitle
	\begin{minipage}{.192\textwidth}
		\centering
		\vskip-10pt
		\includegraphics[width=3.4cm,height=2.5cm]{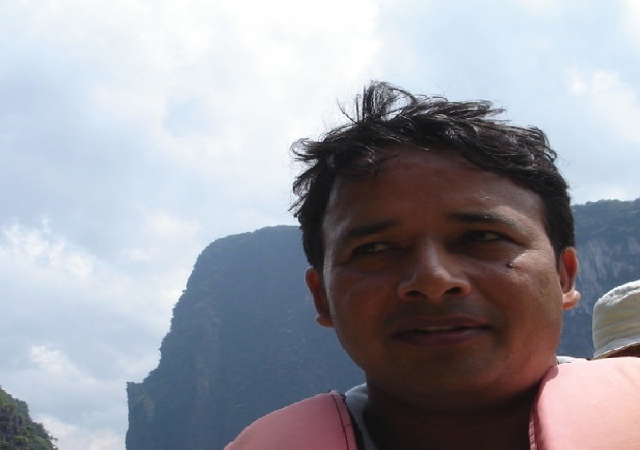}
		\captionsetup{labelformat=empty}
		\captionsetup{justification=centering}
	\end{minipage}
	\begin{minipage}{.192\textwidth}
		\centering
		\vskip-10pt
		\includegraphics[width=3.4cm,height=2.5cm]{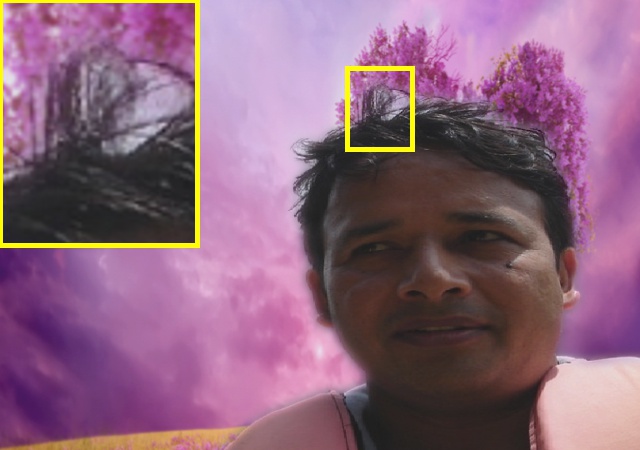}
		\captionsetup{labelformat=empty}
		\captionsetup{justification=centering}
	\end{minipage}
	\begin{minipage}{.192\textwidth}
		\centering
		\vskip-10pt
		\includegraphics[width=3.4cm,height=2.5cm]{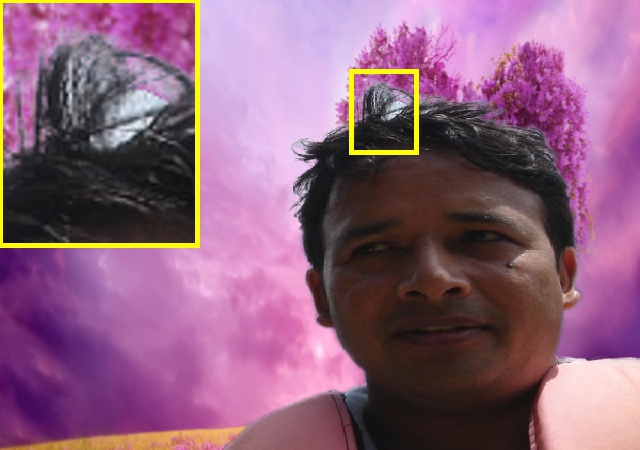}
		\captionsetup{labelformat=empty}
		\captionsetup{justification=centering}
	\end{minipage}
	\begin{minipage}{.192\textwidth}
		\centering
		\vskip-10pt
		\includegraphics[width=3.4cm,height=2.5cm]{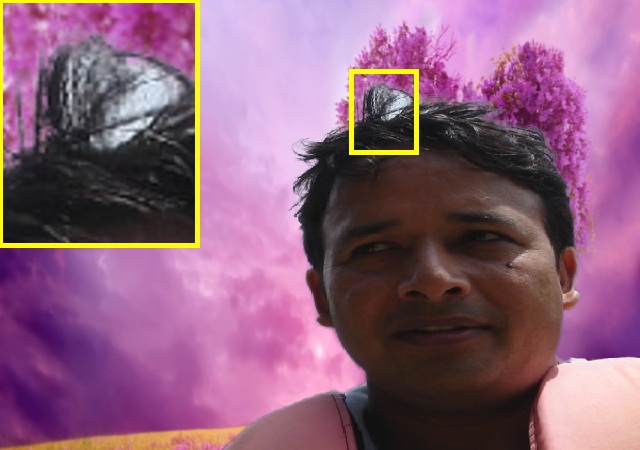}
		\captionsetup{labelformat=empty}
		\captionsetup{justification=centering}
	\end{minipage}
	\begin{minipage}{.192\textwidth}
		\centering
		\vskip-10pt
		\includegraphics[width=3.4cm,height=2.5cm]{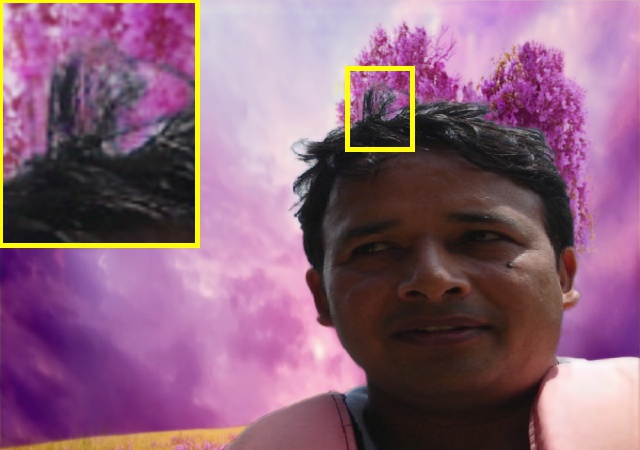}
		\captionsetup{labelformat=empty}
		\captionsetup{justification=centering}
	\end{minipage}
	\vskip+11pt
	\begin{minipage}{.192\textwidth}
		\centering
		\vskip-10pt
		\includegraphics[width=3.4cm,height=3.6cm]{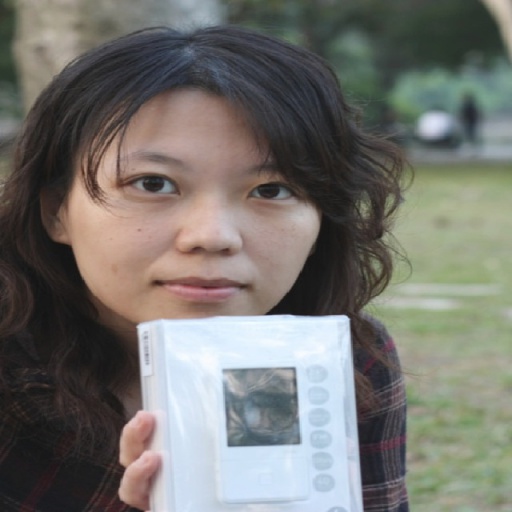}
		\captionsetup{labelformat=empty}
		\captionsetup{justification=centering}
		\vskip-10pt
		\captionof{figure}*{\emph{Forground}}
	\end{minipage}
	\begin{minipage}{.192\textwidth}
		\centering
		\vskip-10pt
		\includegraphics[width=3.4cm,height=3.6cm]{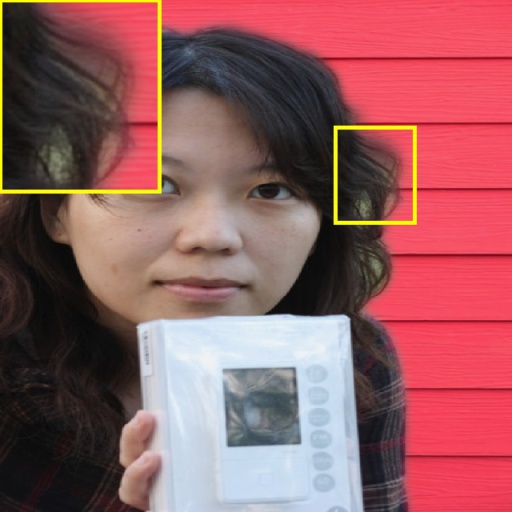}
		\captionsetup{labelformat=empty}
		\captionsetup{justification=centering}
		\vskip-10pt
		\captionof{figure}*{\emph{Lap-pyramid} \cite{blend_laplacian}}
	\end{minipage}
	\begin{minipage}{.192\textwidth}
		\centering
		\vskip-10pt
		\includegraphics[width=3.4cm,height=3.6cm]{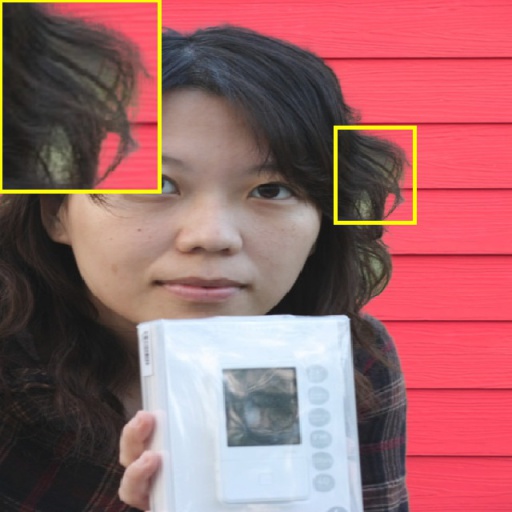}
		\captionsetup{labelformat=empty}
		\captionsetup{justification=centering}
		\vskip-10pt
		\captionof{figure}*{\emph{DIM} \cite{matting_ning}}
	\end{minipage}
	\begin{minipage}{.192\textwidth}
		\centering
		\vskip-10pt
		\includegraphics[width=3.4cm,height=3.6cm]{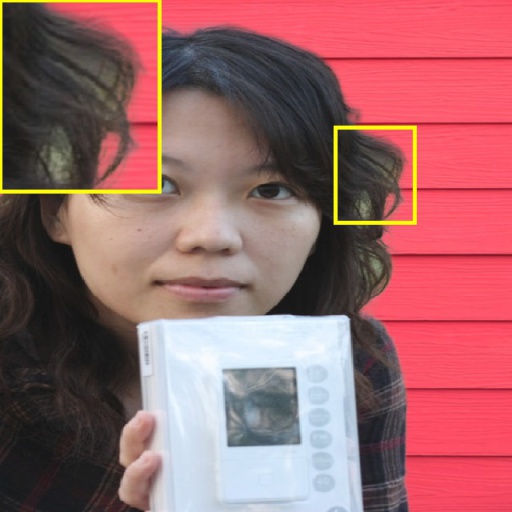}
		\captionsetup{labelformat=empty}
		\captionsetup{justification=centering}
		\vskip-10pt
		\captionof{figure}*{\emph{Index} \cite{matting_index}}
	\end{minipage}
	\begin{minipage}{.192\textwidth}
		\centering
		\vskip-10pt
		\includegraphics[width=3.4cm,height=3.6cm]{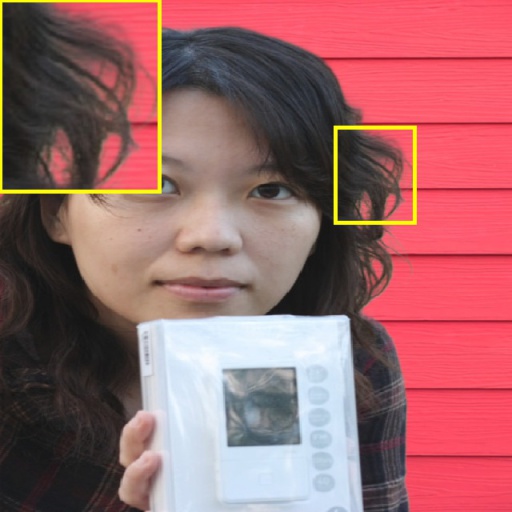}
		\captionsetup{labelformat=empty}
		\captionsetup{justification=centering}
		\vskip-10pt
		\captionof{figure}*{\emph{Our}}
	\end{minipage}
	\vskip-5pt
	\captionof{figure}{Portrait compositing results on the real-world images. All the methods are using the same estimated foreground mask. Previous methods suffer from problems such as halo artifacts and color contamination. Our method learns to generate better compositing results with less boundary artifacts and accurate foreground estimation. Best viewed in color.}\label{fig:teaser}
  \bigskip\bigskip}
\makeatother

\maketitle

\begin{abstract}
Image compositing is a task of combining regions from different images to compose a new image. A common use case is background replacement of portrait images. To obtain high quality composites, professionals typically manually perform multiple editing steps such as segmentation, matting and foreground color decontamination, which is very time consuming even with sophisticated photo editing tools.   
In this paper, we propose a new method which can automatically generate high-quality image compositing without any user input. Our method can be trained end-to-end to optimize exploitation of contextual and color information of both foreground and background images, where the compositing quality is considered in the optimization.  
Specifically, inspired by Laplacian pyramid blending, a dense-connected multi-stream fusion network is proposed to effectively fuse the information from the foreground and background images at different scales. 
In addition, we introduce a self-taught strategy to progressively train from easy to complex cases to mitigate the lack of training data. Experiments show that the proposed method can automatically generate high-quality  composites and outperforms existing methods both qualitatively and quantitatively. 
\end{abstract}

\section{Introduction}

Image compositing is one of the most popular applications in image editing. A common scenario is to composite a portrait photo with a new background. Sample images for portrait compositing is shown in Fig.~\ref{fig:teaser}. To get high-quality composite images, professionals rely on image editing software to perform operations like segmentation, matting and foreground color decontamination. Although many parts of the workflow have been made relatively easier by software, it still requires a lot of expertise and manual efforts to create high-quality composited images. In this paper, we aim to fully automate the portrait image compositing process. 


One straightforward solution is to use a salient object segmentation model \cite{salient1,salient2,salient3,salient4,salient5,salient6} to cut out the foreground region and then paste it on the target background image. However, such simple cut-and-paste approach with the segmentation mask usually results in undesirable artifacts along the object boundary. This is because pixels along the object boundary are usually linear combinations of both foreground and background.  
To address the boundary artifacts, previous approaches resort to low-level image blending methods such as Poisson blending \cite{blend_poisson}, Laplacian blending~\cite{blend_laplacian}, feathering, guided-filter~\cite{blend_guide}, etc. However, these low-level blending methods often introduce other undesirable effects such as color distortion or non-smooth halo artifacts. Sample results are shown in Fig.~\ref{fig:teaser}.   

\begin{figure}
		\centering
		\begin{minipage}{.115\textwidth}
			\centering
			\includegraphics[width=2.05cm,height=2.3cm]{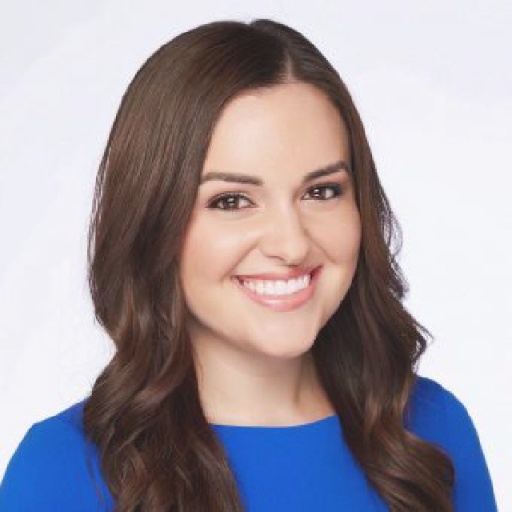}
			\captionsetup{labelformat=empty}
			\captionsetup{justification=centering}
			\vskip-8pt
			\caption*{\emph{Foreground}}
		\end{minipage}
		\begin{minipage}{.115\textwidth}
			\centering
			\includegraphics[width=2.05cm,height=2.3cm]{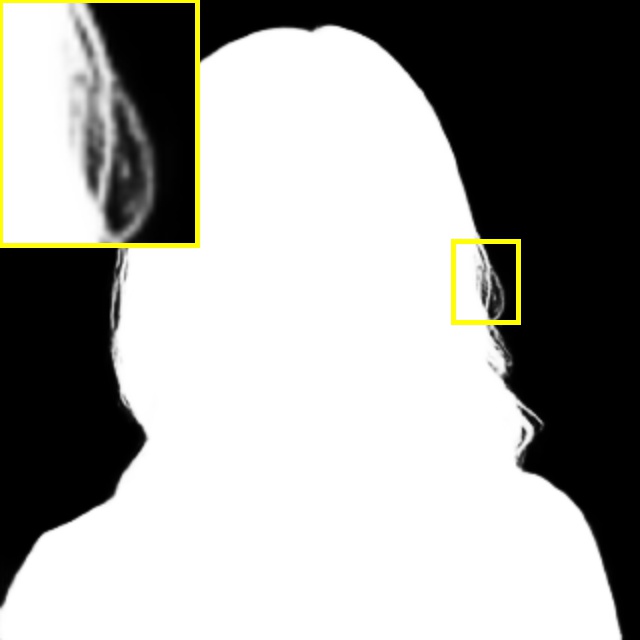}
			\captionsetup{labelformat=empty}
			\captionsetup{justification=centering}
						\vskip-8pt
			\caption*{\emph{GT Mask}}
		\end{minipage}
		\begin{minipage}{.115\textwidth}
			\centering
			\includegraphics[width=2.05cm,height=2.3cm]{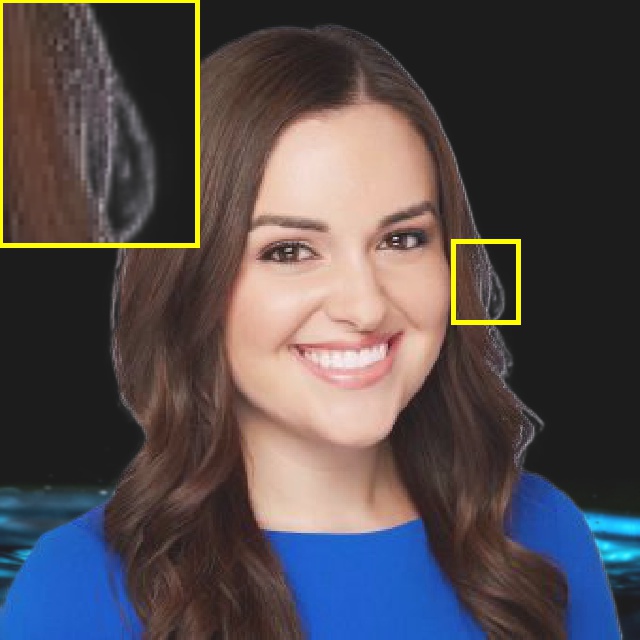}
			\captionsetup{labelformat=empty}
			\captionsetup{justification=centering}
						\vskip-8pt
			\caption*{\emph{Copy-Paste}}
		\end{minipage}
		\begin{minipage}{.115\textwidth}
			\centering
			\includegraphics[width=2.05cm,height=2.3cm]{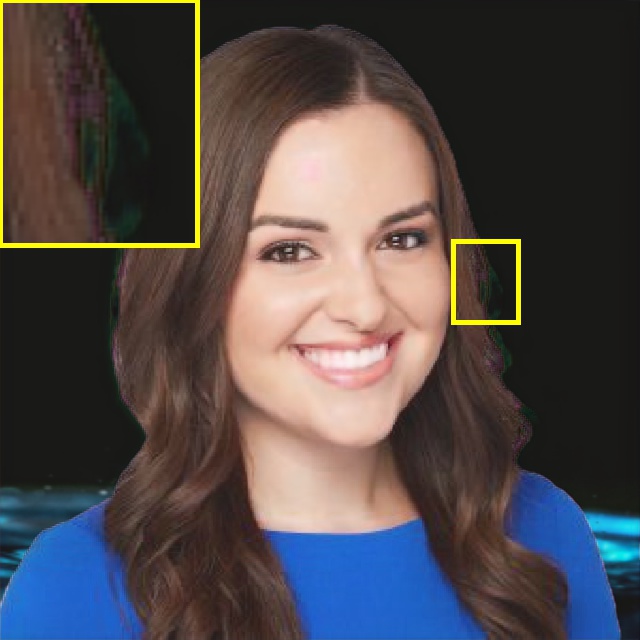}
			\captionsetup{labelformat=empty}
			\captionsetup{justification=centering}
						\vskip-8pt
			\caption*{\emph{Our}}
		\end{minipage}
		\vskip-8pt
		\caption{Leveraging high-quality masks for directly compositing (Copy-paste) may not result in high-quality compositing results. From the direct copy-paste results, it can be clearly observed that color-artifacts are along the boundary.} \label{fig:motivation}
	\vskip-10pt

	\end{figure}

A common solution to the boundary artifacts is to extract the object matte (\emph{i.e.}\ alpha channel) from the foreground image using image matting methods \cite{matting_alibaba,matting_ning,matting_closed_form,matting_ehsan,matting_knn,matting_eccv,matting_xiaoyong, matting_infor}. The ground truth matte controls the linear interpolation of foreground and background in the original input image. Hence, image mattes, if accurately predicted, are able to generate very convincing compositing results with natural blending along the boundary. However, the image matting problem is generally very challenging and it usually requires human input (eg trimap) to identify foreground, background and the uncertain regions to solve. In addition, mistakes in matting are not equally important to image compositing, and the matting methods cannot leverage that as they do not take the end compositing results into consideration. For example, as shown in Fig~\ref{fig:motivation}, though the high-quality ground truth mask (GT mask) is given to cut-out the foreground person and composite it onto a different background, yet the compositing result (Copy-paste result) contains obvious color artifacts along the boundary, which degrade the whole compositing quality. 


In this work, we propose a deep learning based image compositing framework to directly generate a composited portrait image given a pair of foreground and background images. A foreground segmentation network together with a refinement network is employed to extract the portrait mask. Guided by the portrait mask, an end-to-end multi-stream Fusion (MLF) Network is proposed to merge information from both foreground and background images at different scales. The MLF network is inspired by the Laplacian Pyramid Blending method. It uses two encoders to extract different levels of feature maps for the foreground and background images respectively, and fuse them level-by-level through a decoder to reconstruct the final compositing result.  {To notice, the task of the harmonization \cite{harm_sky,harm_tsai,harm_junyan} is different from ours. Their task is to harmonize the appearances (e.g. color) between the foreground and the background and they assume that an artifact-free mask is provided by the user. In contrast, our method is fully automatic and focuses on alleviating the boundary artifacts caused by imperfect foreground masking and color decontamination. Basically, our paper solves an orthogonal problem to color/appearance harmonization for image compositing.}
In addition, we propose an easy-to-hard self-taught based data augmentation scheme to generate high quality compositing data for training the MLF network. The basic idea is to use a MLF network, which is trained on simpler data, to composite more challenging training data for improving itself.

Experimental results evaluated on the synthetic images and real-world images demonstrate the effectiveness of the proposed method compared to previous methods. The superior perceptual quality of our method is validated though a user study.  
Sample results of our method can be found in Figs.~\ref{fig:teaser} and \ref{fig:real} . To summarize, our contributions are 
\begin{itemize} \setlength\itemsep{0em}
	\item an end-to-end deep learning based framework for fully automatic portrait image compositing,
	\item a novel multi-stream Fusion Image Compositing Network for fusing image features at different scales, and
	\item an easy-to-hard data-augmentation scheme for image compositing using self-taught.
\end{itemize}
 \begin{figure*}[t]
	\centering
	\begin{minipage}{0.95\textwidth}
		\centering
		\includegraphics[width=0.85\textwidth]{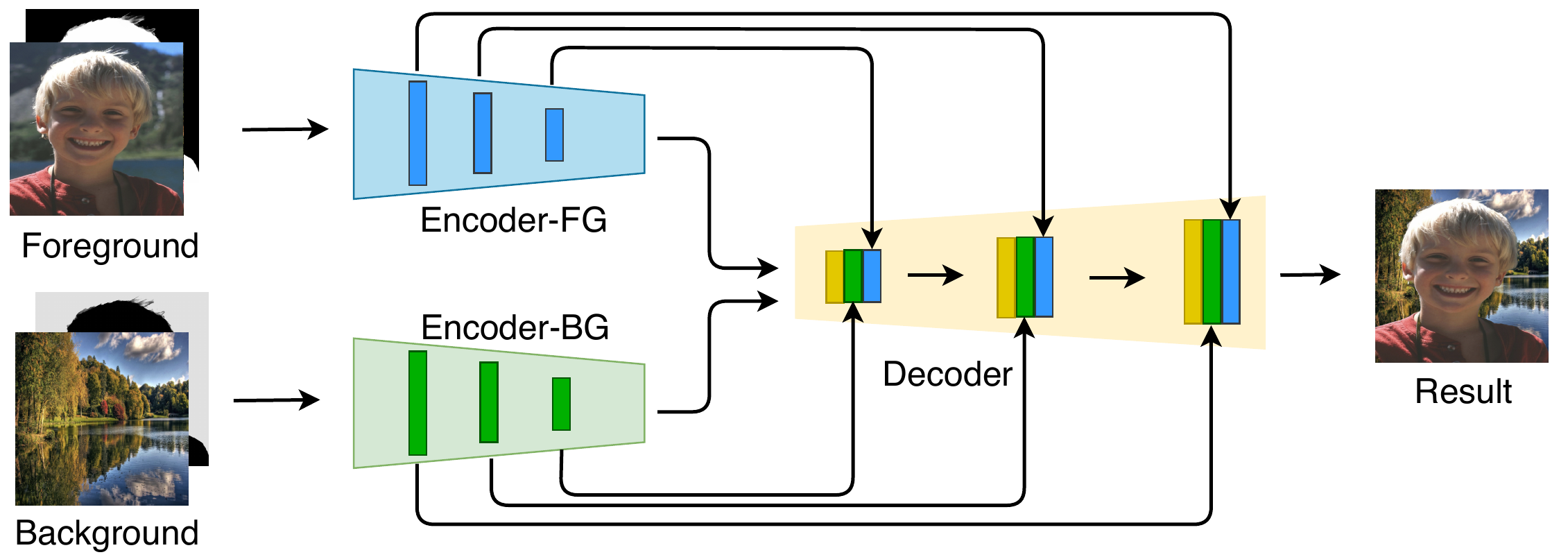}
		\captionsetup{labelformat=empty}
		\captionsetup{justification=centering}
	\end{minipage}
	\vskip-8pt
	\caption{An overview of the proposed multi-stream Fusion Image Compositing Network, where features of different levels are extracted from foreground and background images separately, and then are fused together to generate high-quality compositing results. A pair of masks generated by segmentation models are used to guide the encoding process. See text for more details.}  \label{fig:overview}
\end{figure*}
\section{Related Works}
\subsection{Image Compositing}
Image compositing is a challenging problem in computer vision/ computer graphics, where salient objects from foreground image to be overlayed/composited onto given a background image. And the final goal of the image compositing is to generate realistic high-quality images.   Many image editing applications fall into the category of the image compositing such as image harmonization \cite{harm_tsai,harm_sky, harm_junyan,hard_multi}, image matting \cite{matting_infor,matting_alibaba,matting_knn,matting_xiaoyong,matting_eccv,matting_ehsan,matting_ning,matting_closed_form,matting_survey},   image blending\cite{blend_laplacian,blend_guide,blend_poisson,blending_cvpr}. 

The goal of classical image blending approaches is to guarantee that there is no apparent transition gap between source image and target image.  Alpha blending \cite{blending_alpha} is the simplest and fastest method, but it blurs the fine details and may bring in halo-artifacts in the compositing images. To efficiently leverage the information from different scales,  Burt and Adelson \cite{blend_laplacian} proposed a  multi-scale blending algorithm, named Laplacian pyramid blending. Similar idea has also been used for other low-level tasks such as   image super-resolution \cite{lap_sr} and Generative Adversarial Network (GAN) \cite{lap_gan}.  Alternatively, gradient-based approaches \cite{blend_poisson,blending_po_fast,jia2006drag} also address this problem by adjusting the differences in color and illumination for the composited image globally.  

The most common compositing workflow is based on image matting. Matting refers to the process of extracting the alpha channel foreground object from an image. Traditional matting algorithms \cite{matting_infor,matting_knn,matting_ning,matting_ehsan,matting_index,matting_disentangled} require a user defined trimap which limits their applications in automatic image process. Though recent works \cite{matting_xiaoyong,matting_alibaba} have leveraged the CNN models to automatically  generate trimaps, they still regard trimap generation and alpha channel computation as two separated stages. In addition,  the matting methods  do not take the final compositing results into consideration. Instead, we propose an end-to-end image compositing framework that takes the final compositing performance as the optimization objective. 
\subsection{Data Augmentation}
Data augmentation is a common technique to improve the training of deep neural networks. It helps to reduce over-fitting and improves the model generalization. Data-augmentation has been successfully applied to various computer vision applications both in low-level vision and high-level vision such as image matting \cite{matting_ning, matting_alibaba}, image harmonization~\cite{harm_tsai}, and object detection \cite{data-aug1}. Most data augmentation methods are based on trivial transformations such as cropping, flipping, color shift, or adding noise to an image \cite{data-aug1,harm_tsai, yizhu_seg, syn_seg1}. In our problem, the training requires a triplet composed of a pair of foreground and background images and the target composited images, and traditional data augmentation methods cannot help to diversify the contents of the triplet. In this paper, we propose a self-taught method to automatically generate such triplet samples for our image compositing problem.

\section{Deep Image Compositing}

In this section, we present our Deep Image Compositing framework. Although we only implement it for portrait compositing in this paper, the formulation of the framework is general and we hope it can be useful in other image compositing applications, too.

The proposed framework takes a pair of foreground and background images as input, and generates the composited image.
It has three components: 1) a Foreground Segmentation Network, 2) a Mask Refinement Network and 3) a multi-stream Fusion Network. First, the segmentation network automatically extracts an object mask from the foreground image. Then, the mask refinement network takes the image and the mask as input to refine the mask boundary. After that, the refined mask, together with the foreground and background images, is passed to the multi-stream fusion network to generate the compositing results. We will describe these components as follows.

\subsection{Multi-stream Fusion Network for Compositing}

We present the multi-stream Fusion (MLF) Network first as it is independent of the other two components and can work with other segmentation and matting models, too. The goal of the MLF network is to provide natural blending between the foreground cutout and the background image, removing artifacts caused by color contamination, aliasing and the inaccuracy of the segmentation mask. 

Our MLF network is inspired by the Laplacian pyramid method for image blending. The Laplacian blending \cite{blend_laplacian} method computes image pyramid representations for both foreground and background images, and then blends different levels of details with varied softness along the mask boundary through the image pyramid representations. The final composited image can be reconstructed from the multi-stream fused image representations.

Similarly, the proposed MLF network consists of two separate encoders to extract the multi-scale features from foreground and background images separately. The input to both encoders is a concatenation of the image and a pre-computed soft mask. The mask for the foreground image is computed by our segmentation and refinement networks, and the mask for the background image is an inverted version of it. The foreground and background encoders then generate different levels of features, which correspond to the image pyramid representations in Laplacian blending.

At the end of the encoders, the highest-level of feature maps are concatenated and are passed to a decoder. At different decoding levels, feature maps get upsampled through deconvolution and are concatenated with the same level feature maps from the two encoders. At the end of the decoder, the composited image is reconstructed from the fused feature maps. This process is analogous to the fusion and reconstruction process of Laplacian pyramid blending. Dense-block \cite{dense_net} is leveraged as the basic building block for the discussed encoder and decoder architecture. Details of the proposed network is presented in the supplementary material. The overview of the MLF network is shown in Fig.~\ref{fig:overview}.

As we can see, our MLF network can be regarded as an extension of the popular encoder-decoder network with short connections \cite{u-net,badrinarayanan2017segnet}. Instead of a single-stream encoder in the previous models, our two-stream encoder pipeline encodes the foreground and background feature maps separately, which are fused later during the decoding process. We find that such two-stream design not only coincides with the Laplacian blending framework, but also provides better performance than the single-stream design in our experiment.

For training, the proposed method is optimized via both L1 loss and perceptual loss \cite{perceptual_loss,ledig2017photo} to encourage image-level perceptual realism \cite{zhang2018unreasonable} for the composited images.  The loss function is defined as follows: 
\begin{equation}
\label{eq:overall_loss}
L_{\text{all}} = L_{1} + \lambda_PL_{P},
\end{equation}
where $L_{1}$ denotes the L1  loss and  $L_P$ indicates the perceptual loss. Here, $\lambda_P$ indicates the weights for the perceptual loss. The perceptual loss is evaluated on relu 1-1 and relu 2-1 layers of the pre-trained VGG \cite{vggface} model.


\subsection{Segmentation and Mask Refinement Networks}

	\begin{figure}
		\centering
		\begin{minipage}{.15\textwidth}
			\centering
			\includegraphics[width=2.6cm,height=2.8cm]{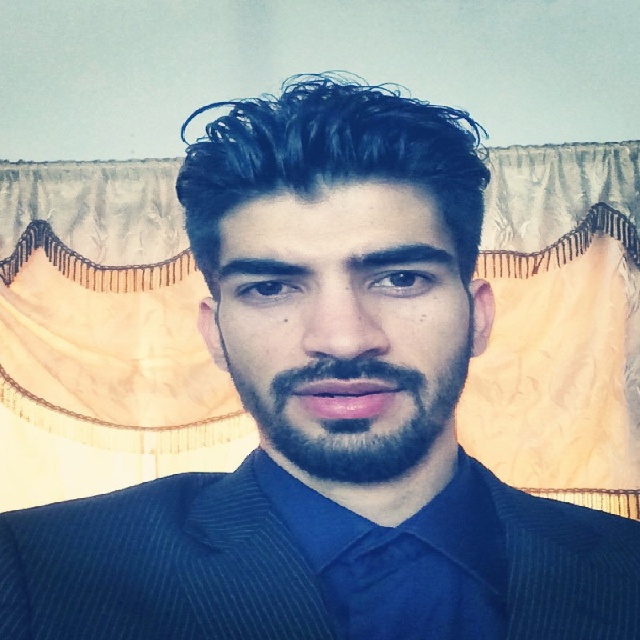}
			\captionsetup{labelformat=empty}
			\captionsetup{justification=centering}
			\vskip-8pt
			\caption*{\emph{Foreground}}
		\end{minipage}
		\begin{minipage}{.15\textwidth}
			\centering
			\includegraphics[width=2.6cm,height=2.8cm]{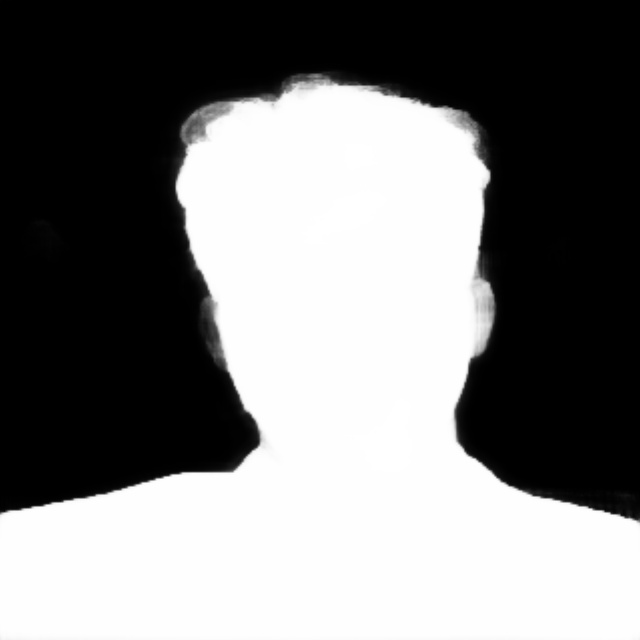}
			\captionsetup{labelformat=empty}
			\captionsetup{justification=centering}
						\vskip-8pt
			\caption*{\emph{Raw Mask}}
		\end{minipage}
		\begin{minipage}{.15\textwidth}
			\centering
			\includegraphics[width=2.6cm,height=2.8cm]{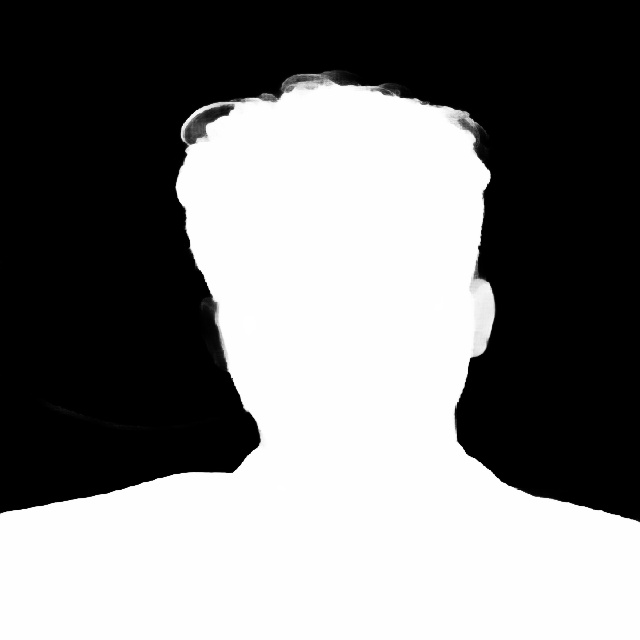}
			\captionsetup{labelformat=empty}
			\captionsetup{justification=centering}
						\vskip-8pt
			\caption*{\emph{Refined Mask}}
		\end{minipage}
		\vskip-8pt
		\caption{Results on the  segmentation mask estimation. It can be observed that the refined segmentation mask preserves better boundary details with more confidence.} \label{fig:mask}
			\vskip-10pt

	\end{figure}

The foreground segmentation network can be implemented as a salient object segmentation model \cite{salient1,salient2,salient3,salient4,salient5,salient6} or specifically a portrait segmentation model \cite{matting_xiaoyong, matting_alibaba,shen2016automatic}, and we refer the readers to those related works for the details of model training and datasets. In our implementation, we use a salient object segmentation model \cite{ano} due to its speed and accuracy, but that our framework can work with any off-the-shelf salient object segmentation models.

However, the raw mask from the foreground segmentation model is often not very accurate at the object boundaries. The segmentation network also processes the image at low resolution, so the upsampled mask will further suffer from the up-sampling artifacts like jagged boundaries. Therefore, we propose a mask refinement network to refine the details along the object boundary and up-sample the mask with fewer artifacts.

The refinement network shares the same architecture as the segmentation network, except that input is a four-channel RGB-A image, where the fourth contains the raw segmentation mask. To make this mask refinement network focus on different levels of local details, we sample image patches of various sizes for the training. In training, a cropped version of the image and the pre-computed raw mask are passed to the refinement network to generate the local refined mask. The training uses the same data and and the same cross-entropy loss as used by the segmentation model.
At testing, the refinement network takes the whole image and its mask as input. 

The refinement network can be applied recursively at different scales. In our implementation, we first resize the image and the its raw mask to $320\times320$ and generate a refined mask at this resolution. Then we resize the image to $640\times640$ and upsample the refinemask to same size, and apply the refinement network again. We find this two-stage refinement scheme working very well in practice. A sample result of the refinement network is shown in Fig.~\ref{fig:mask}. It can be observed that the refined mask preserves much better boundary details and reduce the fuzziness of the raw mask. This also makes the training of our fusion network easier.

\section{Easy-to-Hard Data Augmentation}\label{sec:data}

To train our multi-stream Fusion (MLF) network, each training sample is a triplet $[FG, BG, C]$, where $FG$ is the foreground image, $BG$ is the background image and $C$ is the target composted image of $FG$ and $BG$. As we want the MLF network to learn to produce a visually pleasing blending between $FG$ and $BG$, the quality of the target image $C$ is the key to our method. However, manually creating such high-quality compositing dataset requires expert-level human effort, which limits the scalability of the training data collection.

	\begin{figure}[t]
		\centering
		\begin{minipage}{0.5\textwidth}
			\centering
			\includegraphics[width=1\textwidth]{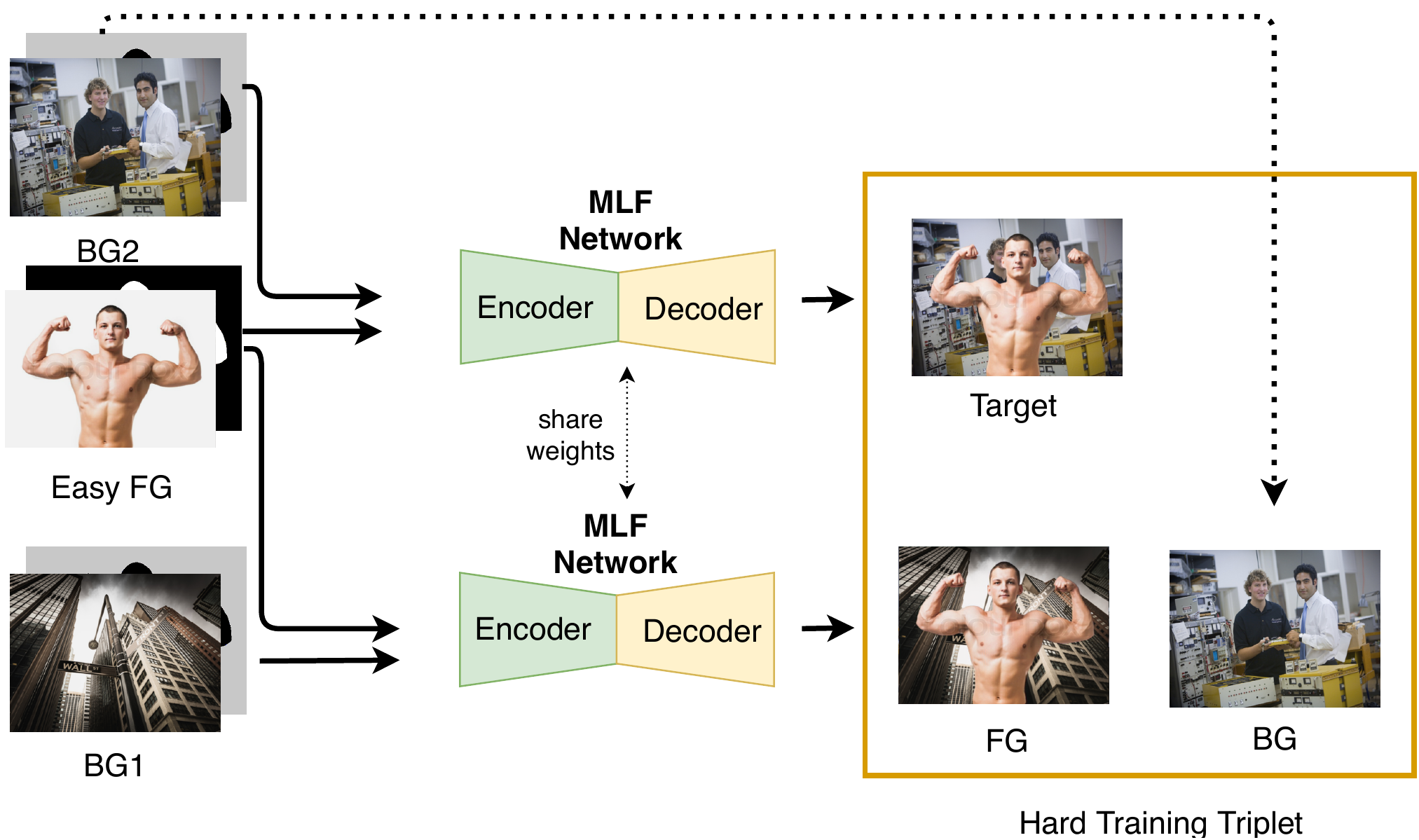}
			\captionsetup{labelformat=empty}
			\captionsetup{justification=centering}
		\end{minipage}
		\vskip-8pt
		\caption{An overview of the proposed easy-to-hard data augmentation procedure.} \label{fig:data_augmen}
	\vskip-8pt
	\end{figure}
	
To address this issue and generate a relative large-scale image compositing dataset without much human annotation effort, we propose an easy-to-hard data-augmentation approach using a self-taught scheme. The basic idea is to use the MLF network to generate more challenging data to improve itself. The MLF network is first trained on a few easy training triplets where the foreground images $FG$ are all portrait images with simple color background. After that, we collect a lot of such simple portrait images and use the MLF network to generate more challenging training triplets for the next training stage. The overview of this data augmentation scheme is shown in Fig.~\ref{fig:data_augmen} and we describe more details below.
	 
	
We first use a small matting dataset \cite{matting_ning} to create a simple compositing training set. Images in the matting dataset have alpha channel and were processed by color decontamination. Thus, they can be composited to any background images using the alpha channel. To generate a easy training triplet, the foreground images $FG$ is generated by compositing the matting image with pure color background; the background image $BG$ can be a random web image; and the target image $C$ is created by using the alpha channel of the matting image as well. By firstly training our MLF network on these triplets, the network learns to blend an easy foregrond image onto a random background image.  

We then use our specifically trained MLF network to generate harder training triplets. We collect a lot of web portrait images with simple background, with which we generate composited images with random background images using the MLF network (see Fig.~\ref{fig:data_augmen}). Given a simple portrait image, which is denoted as Easy $FG$ in Fig.~\ref{fig:data_augmen}, we sample two random background images $BG1$ and $BG2$ to generate two composited images using the MLF network. Without loss of generality, the composited image of $BG2$ and Easy $FG$ is used as a new target image $C'$; the other composited image then becomes a new foreground image $FG'$, and $BG2$ will be used as the new background image $BG'$. As we can see from Fig.~\ref{fig:data_augmen}, the new triplet $[FG', BG', C']$ follows the compositing relationship
\begin{equation}
    C' = \text{Easy }FG \oplus BG2 = FG' \oplus BG',
\end{equation}
whtere Easy $FG$ and $FG'$ share the same foreground content, and $BG'=BG2$. The compositing operation is denoted as $\oplus$.

\begin{figure}
		\centering


		\begin{minipage}{.15\textwidth}
			\centering
			\includegraphics[width=2.6cm,height=2.8cm]{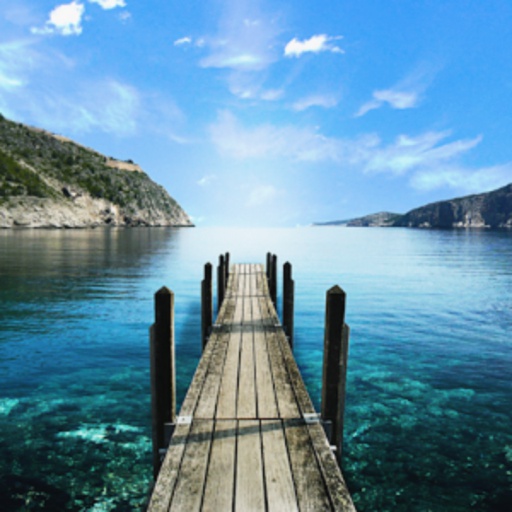}
			\captionsetup{labelformat=empty}
			\captionsetup{justification=centering}
			\vskip-8pt
			\caption*{\emph{Background}}
		\end{minipage}
		\begin{minipage}{.15\textwidth}
			\centering
			\includegraphics[width=2.6cm,height=2.8cm]{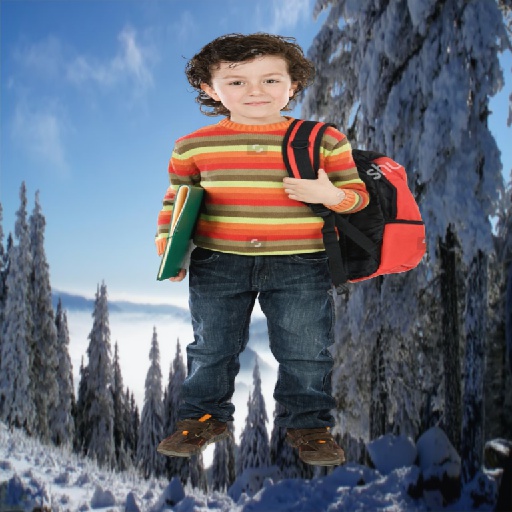}
			\captionsetup{labelformat=empty}
			\captionsetup{justification=centering}
			\vskip-8pt
			\caption*{\emph{Foreground}}
		\end{minipage}
		\begin{minipage}{.15\textwidth}
			\centering
			\includegraphics[width=2.6cm,height=2.8cm]{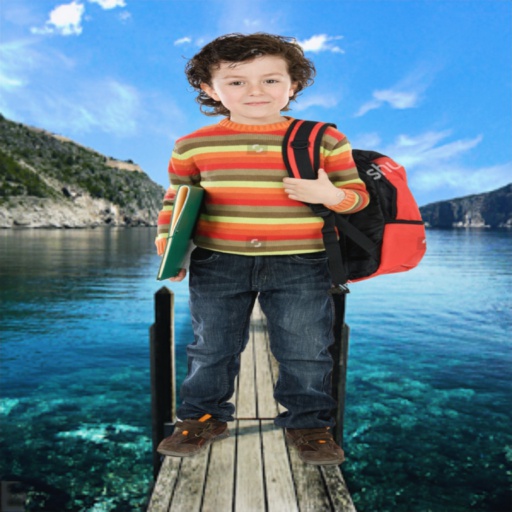}
			\captionsetup{labelformat=empty}
			\captionsetup{justification=centering}
			\vskip-8pt
			\caption*{\emph{Target}}
		\end{minipage}
		\vskip-8pt
		\caption{Samples of triplets generated by our self-taught data augmentation algorithm. It can be observed that the proposed data augmentation algorithm is able to generate high-quality with near-perfect target images.  } \label{fig:triplet}
			\vskip-8pt

	\end{figure}

In this way, we generated high-quality hard triplets to augment the original matting training set. Sample triplets are shown in Fig.~\ref{fig:triplet}. It can be observed that the proposed data augmentation algorithm is able to generate high-quality compositing targets. 	
Our results in the next section show that these self-generated training sample is essential for the good performance of our method.  

\section{Experiments}\label{sec:exper}
\begin{figure*}[h]
			\vskip-15pt
		\centering
		\begin{minipage}{.182\textwidth}
			\centering
			\includegraphics[width=3.0cm,height=3.2cm]{results//937_fore.jpg}
			\captionsetup{labelformat=empty}
			\captionsetup{justification=centering}
						\vskip-7pt
			\caption*{\emph{Foreground}}
		\end{minipage}
		\begin{minipage}{.182\textwidth}
			\centering
			\includegraphics[width=3.0cm,height=3.2cm]{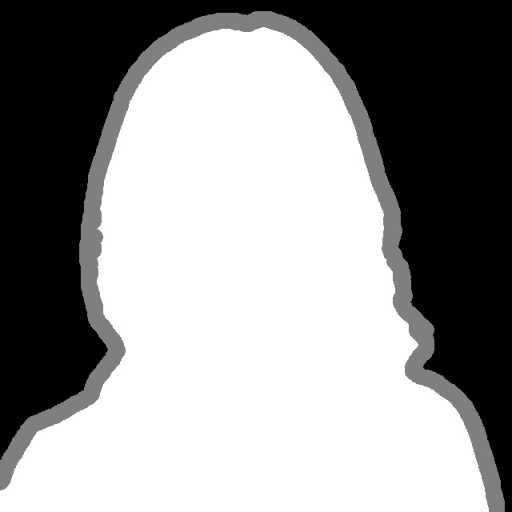}
			\captionsetup{labelformat=empty}
			\captionsetup{justification=centering}
						\vskip-7pt
			\caption*{\emph{Generated  Trimap}}
		\end{minipage}
		\begin{minipage}{.182\textwidth}
			\centering
			\includegraphics[width=3.0cm,height=3.2cm]{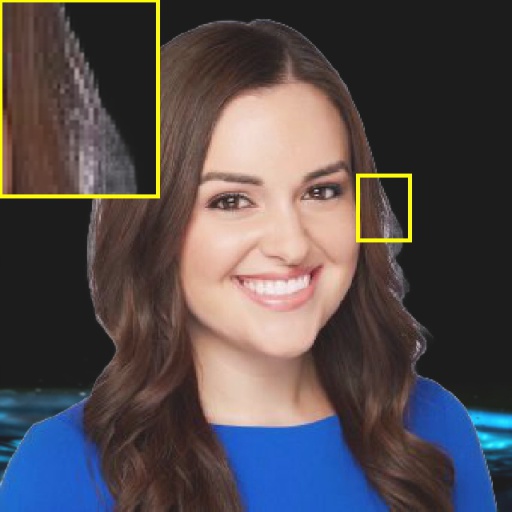}
			\captionsetup{labelformat=empty}
			\captionsetup{justification=centering}
						\vskip-7pt
			\caption*{\emph{Copy-Paste}}
		\end{minipage}
		\begin{minipage}{.182\textwidth}
			\centering
			\includegraphics[width=3.0cm,height=3.2cm]{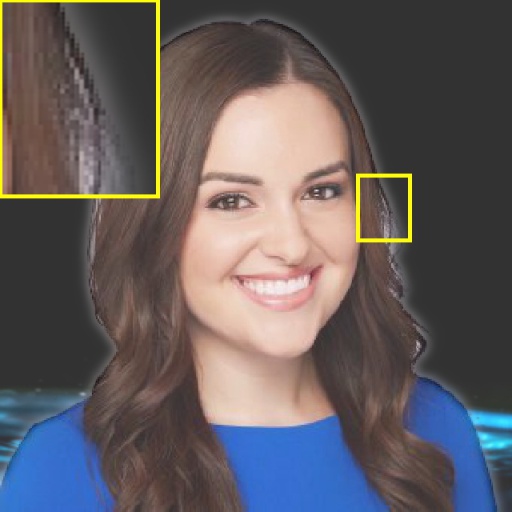}
			\captionsetup{labelformat=empty}
			\captionsetup{justification=centering}
									\vskip-7pt
			\caption*{\emph{Lap-Pyramid} \cite{blend_laplacian}}
		\end{minipage}	
		\begin{minipage}{.182\textwidth}
			\centering
			\includegraphics[width=3.0cm,height=3.2cm]{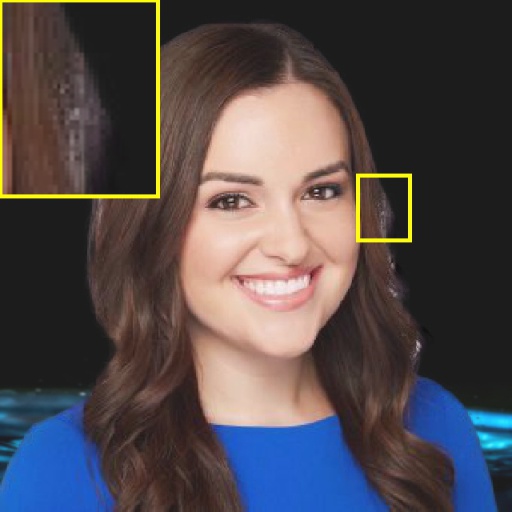}
			\captionsetup{labelformat=empty}
			\captionsetup{justification=centering}
									\vskip-7pt
			\caption*{\emph{Closed} \cite{matting_closed_form}}
		\end{minipage}
		\\		\vskip+11pt

		\begin{minipage}{.182\textwidth}
			\centering
			\vskip-8pt
			\includegraphics[width=3.0cm,height=3.2cm]{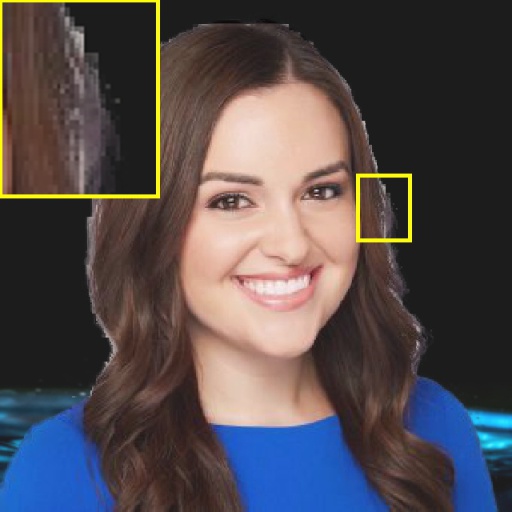}
			\captionsetup{labelformat=empty}
			\captionsetup{justification=centering}
									\vskip-7pt
			\caption*{\emph{KNN} \cite{matting_knn}}
		\end{minipage}
		\begin{minipage}{.182\textwidth}
			\centering
			\vskip-8pt
			\includegraphics[width=3.0cm,height=3.2cm]{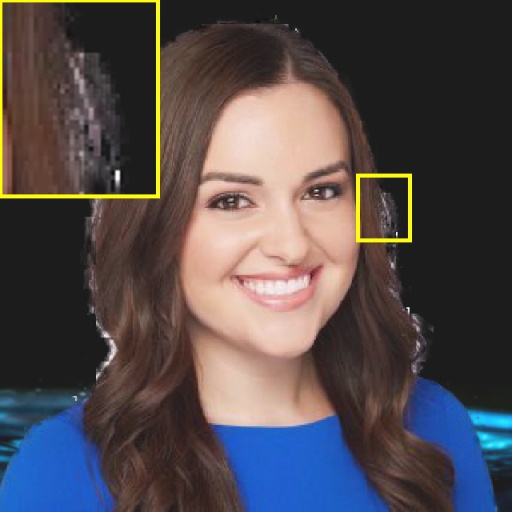}
			\captionsetup{labelformat=empty}
			\captionsetup{justification=centering}
									\vskip-7pt
			\caption*{\emph{Info-flow \cite{matting_infor}}}
		\end{minipage}	
		\begin{minipage}{.182\textwidth}
			\centering
			\vskip-8pt

			\includegraphics[width=3.0cm,height=3.2cm]{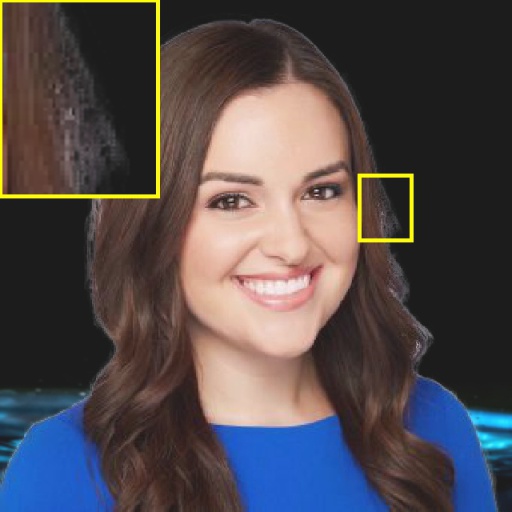}
			\captionsetup{labelformat=empty}
			\captionsetup{justification=centering}
						\vskip-7pt
			\caption*{\emph{DIM \cite{matting_ning}}}
		\end{minipage}
		\begin{minipage}{.182\textwidth}
			\centering
				\vskip-8pt

			\includegraphics[width=3.0cm,height=3.2cm]{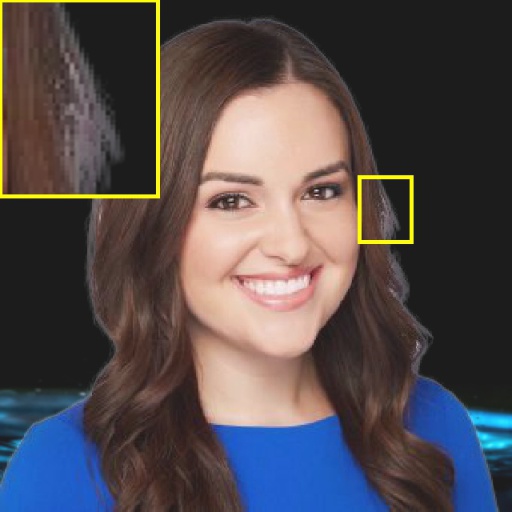}
			\captionsetup{labelformat=empty}
			\captionsetup{justification=centering}
						\vskip-7pt
			\caption*{\emph{Index \cite{matting_index}}}
		\end{minipage}
		\begin{minipage}{.182\textwidth}
			\centering
			\vskip-8pt
			\includegraphics[width=3.0cm,height=3.2cm]{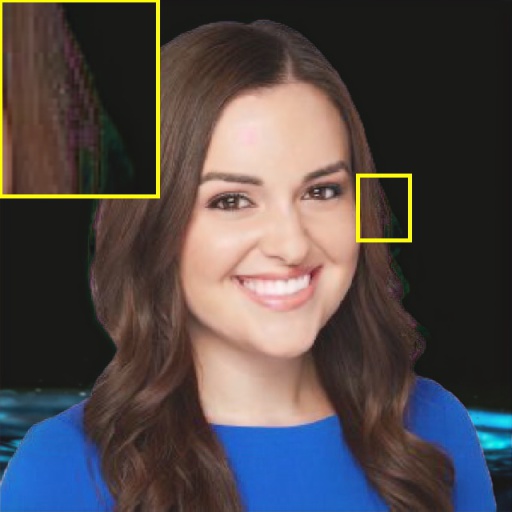}
			\captionsetup{labelformat=empty}
			\captionsetup{justification=centering}
									\vskip-7pt
			\caption*{\emph{Our}}
		\end{minipage}
	\\			\centering (Image 2)	\vskip+2pt 		
	\begin{minipage}{.182\textwidth}
		\centering
		\includegraphics[width=3.0cm,height=2.9cm]{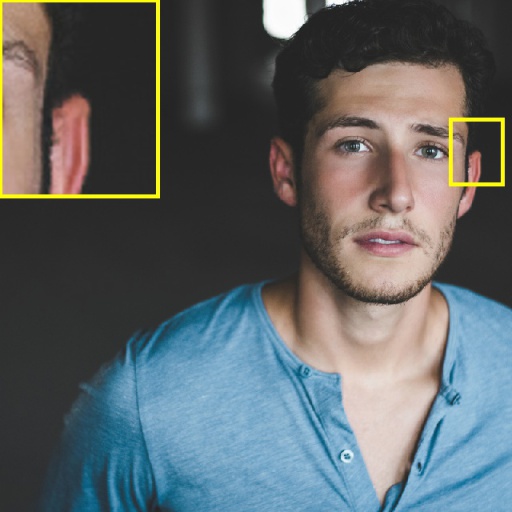}
		\captionsetup{labelformat=empty}
		\captionsetup{justification=centering}
				\vskip-7pt
		\caption*{\emph{Foreground}}
	\end{minipage}
	\begin{minipage}{.182\textwidth}
		\centering
		\includegraphics[width=3.0cm,height=2.9cm]{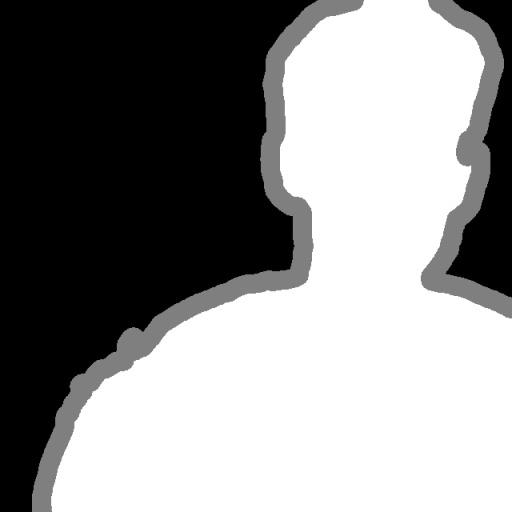}
		\captionsetup{labelformat=empty}
		\captionsetup{justification=centering}
						\vskip-7pt
		\caption*{\emph{Generated  Trimap}}
	\end{minipage}
	\begin{minipage}{.182\textwidth}
		\centering
		\includegraphics[width=3.0cm,height=2.9cm]{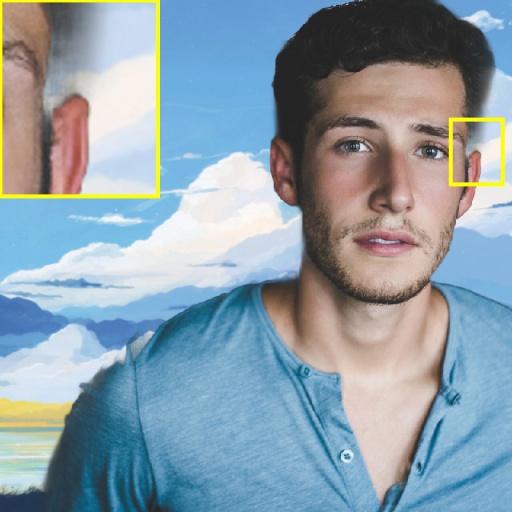}
		\captionsetup{labelformat=empty}
		\captionsetup{justification=centering}
						\vskip-7pt
		\caption*{\emph{Copy-Paste}}
	\end{minipage}
	\begin{minipage}{.182\textwidth}
		\centering
		\includegraphics[width=3.0cm,height=2.9cm]{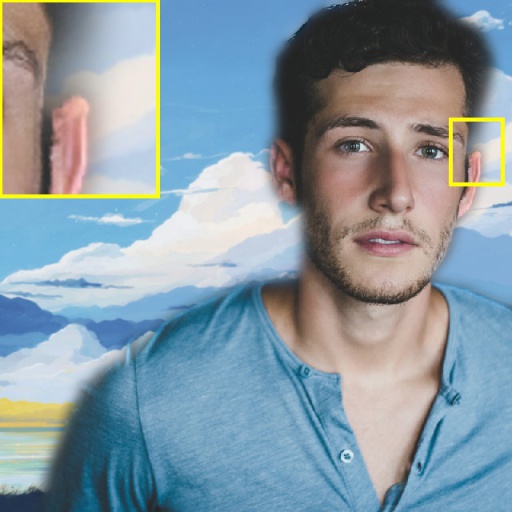}
		\captionsetup{labelformat=empty}
		\captionsetup{justification=centering}
						\vskip-7pt
		\caption*{\emph{Lap-Pyramid \cite{blend_laplacian}}}
	\end{minipage}
	\begin{minipage}{.182\textwidth}
		\centering
		\includegraphics[width=3.0cm,height=2.9cm]{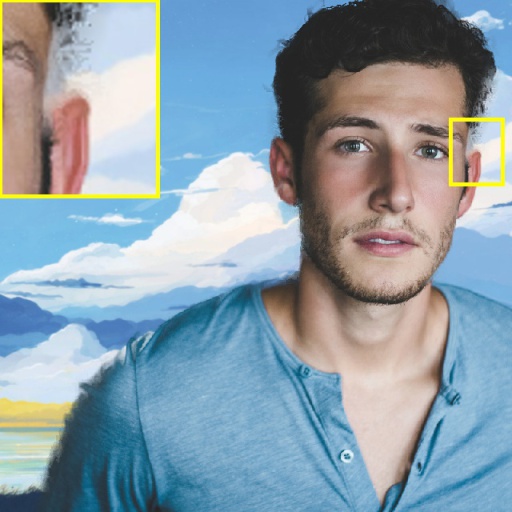}
		\captionsetup{labelformat=empty}
		\captionsetup{justification=centering}
								\vskip-7pt
		\caption*{\emph{Closed} \cite{matting_closed_form}}
	\end{minipage}
	\\		\vskip+11pt
	\begin{minipage}{.182\textwidth}
		\centering
												\vskip-8pt
		\includegraphics[width=3.0cm,height=2.9cm]{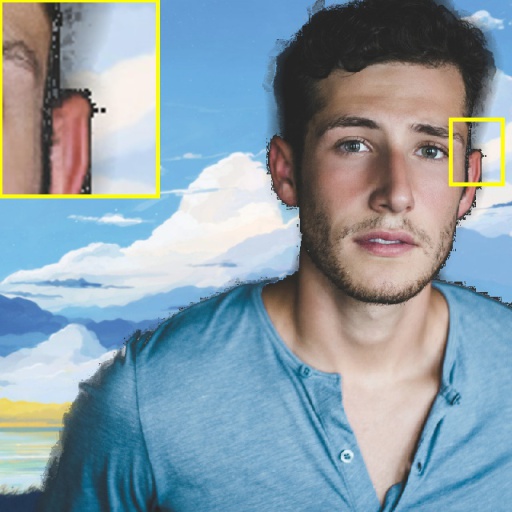}
		\captionsetup{labelformat=empty}
		\captionsetup{justification=centering}
								\vskip-7pt
		\caption*{\emph{KNN} \cite{matting_knn}}
	\end{minipage}
	\begin{minipage}{.182\textwidth}
		\centering
												\vskip-8pt
		\includegraphics[width=3.0cm,height=2.9cm]{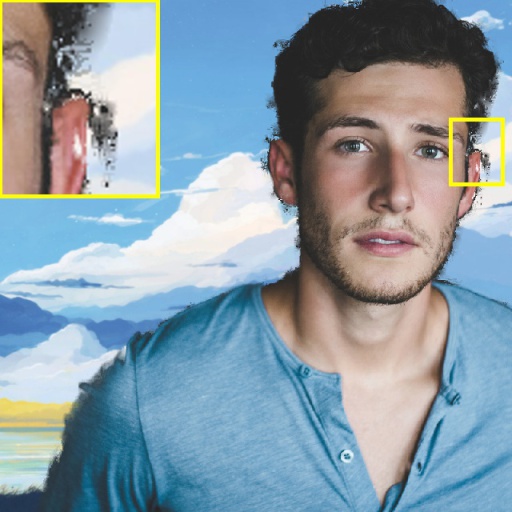}
		\captionsetup{labelformat=empty}
		\captionsetup{justification=centering}
								\vskip-7pt
		\caption*{\emph{Info-flow} \cite{matting_infor}}
	\end{minipage}	
	\begin{minipage}{.182\textwidth}
		\centering
		\includegraphics[width=3.0cm,height=2.9cm]{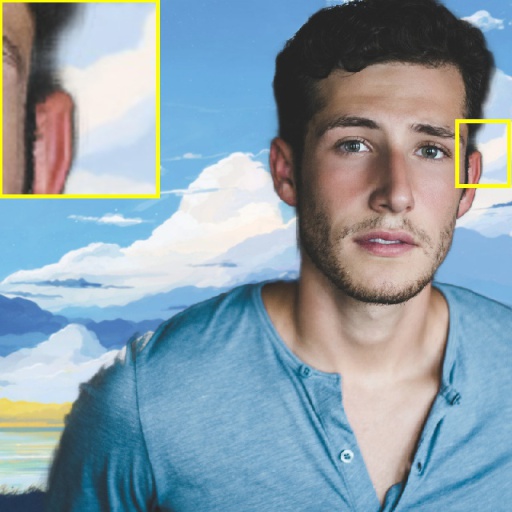}
		\captionsetup{labelformat=empty}
		\captionsetup{justification=centering}
						\vskip-7pt
		\caption*{\emph{DIM \cite{matting_ning}}}
	\end{minipage}
	\begin{minipage}{.182\textwidth}
		\centering
		\includegraphics[width=3.0cm,height=2.9cm]{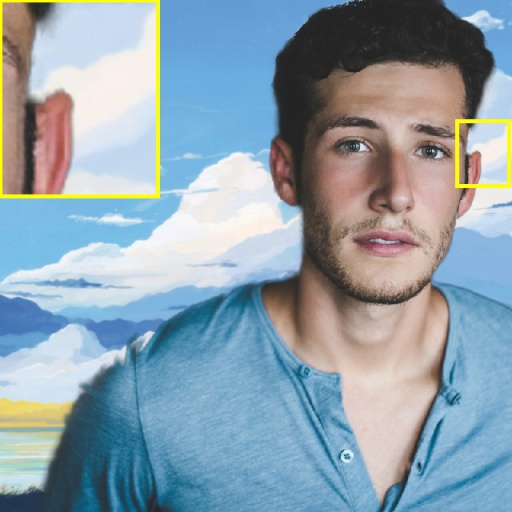}
		\captionsetup{labelformat=empty}
		\captionsetup{justification=centering}
						\vskip-7pt
		\caption*{\emph{Index \cite{matting_index}}}
	\end{minipage}	
	\begin{minipage}{.182\textwidth}
		\centering
												\vskip-8pt
		\includegraphics[width=3.0cm,height=2.9cm]{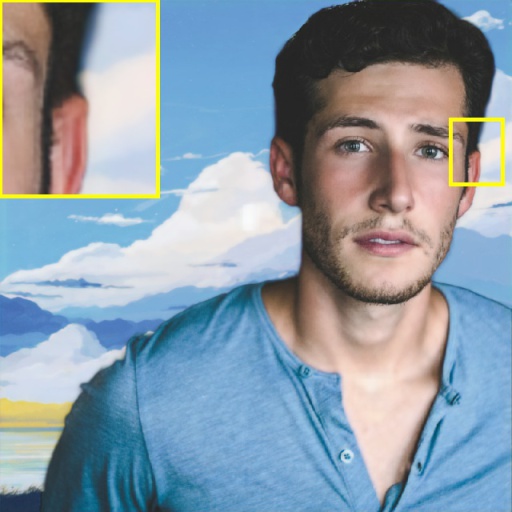}
		\captionsetup{labelformat=empty}
		\captionsetup{justification=centering}
														\vskip-7pt
		\caption*{\emph{Our}}
	\end{minipage}
	\\			\centering (Image 3)	\vskip+2pt 		
	\vskip-8pt
	\caption{Results compared with other methods.} \label{fig:real}
\end{figure*}

\begin{table*}
	\centering
	\small
	\vskip-6pt

	\caption{Quantitative results evaluated on \textbf{SynTest} compared with the other  methods. Evaluation on unknown regions.}
			\vskip-6pt

	\begin{tabular}{|c|cc|ccccc|c|}
		\hline
		& Copy-paste   & Lap-Pyramid \cite{blend_laplacian} & Closed \cite{matting_closed_form} &KNN \cite{matting_knn} & Info-flow \cite{matting_infor} & DIM  \cite{matting_ning} & Index \cite{matting_index} & Our\\ \hline\hline
		PSNR (dB) &  17.88  &  17.85 &  18.23  &  18.21 & {17.88 }& 18.40  & {19.01} & \textbf{19.34} \\ \hline
	\end{tabular}
	\vskip-6pt

	\label{tab:state}
\end{table*}

We evaluate our deep image compositing method through quantitative and qualitative evaluations. A user study is also performed to measure the users' preference regarding the perceptual quality of the compositing results. Finally, we perform some ablation studies to validate the main components of our method. 

\textbf{Datasets:}
The segmentation and refinement network is trained on the DUTS \cite{wang2016saliency,yang2013saliency}, MSRA-10K \cite{HouPami19Dss,HouCvpr2017Dss} and Portrait segmentation \cite{matting_xiaoyong,shen2016automatic} datasets .
The multi-stream fusion compositing network is trained using the synthesized dataset via the proposed self-taught data augmentation method together with a matting-based compositing dataset. Similar to \cite{matting_ning}, the matting-based compositing dataset is composed of 30000 training images generated by compositing foreground images to random background images using the ground truth mattes. In addition, we also synthesize a testing dataset using the proposed self-taught data-augmentation method, denoted as \textbf{SynTest}. SynTest is used for quantitative evaluations. We leverage the PSNR as the measurements to measure the final compositing quality. 

\textbf{Implementation Details:}
The segmentation and refinement module is optimized via ADAM algorithm with a learning rate of $2 \times 10^{-3}$ and a batch size of 8. All the training samples for the segmentation and refinement modules are resized to 256 $\times$ 256.

Similarly, the multi-stream fusion compositing network is trained using  ADAM with a learning rate of $2 \times 10^{-3}$ and batch size of 1. All the training samples are cropped and resized to 384 $\times$ 384 and all the testing samples are resized to $768 \times 768$. We trained the network for 200000 iterations and  choose $\lambda_P=0.8$ for the loss in Eqn.~\ref{eq:overall_loss}.  Details of the proposed multi-stream fusion network is discussed in the supplementary material.

\textbf{Compared Methods:}
In this paper, the proposed method is compared with traditional blending-based compositing methods such as Laplacian pyramid blending \cite{blend_laplacian}. We also evaluate the matting-based compositing approach using state-of-the-art matting methods such as Closed-Form (Closed) \cite{matting_closed_form}, KNN \cite{matting_knn}, Information-Flow (Info-flow) \cite{matting_infor}, Deep Image Matting (DIM) \cite{matting_ning} and Index-net \cite{matting_index}. In addition, we also compare one baseline method called copy-paste. For copy-paste, the refined segmentation mask estimated from the refinement segmentation module is used as the soft alpha matte for the compositing.  

For fair comparison, all the compared methods use the same refined mask as our method. For the feathering method, we apply Gaussian blur with $\sigma = 2$ to soften the mask.   For the Laplacian pyramid blending \cite{blend_laplacian}, we use the OpenCV implementation. 
As matting-based methods require trimaps, we binarize the refined mask and then generate a pseudo-trimap by labeling a narrow boundary band of width 16 as unknown. Sample trimaps are shown in Fig.~\ref{fig:real}.  To notice, an automatic color-decontamination algorithm \cite{matting_closed_form} is used for the matting-based compositing methods to enhance their compositing quality. 

\begin{table*}
	\centering
	\small
	\vskip-12pt

    \caption{User-study results compared with the other methods. (Lower is the better.)}
	\vskip-6pt
		\begin{tabular}{|c|ccccc|cc|c|}
			\hline
			 & Lap-Pyramid \cite{blend_laplacian} & Closed \cite{matting_closed_form} & Info-flow \cite{matting_infor} & DIM \cite{matting_ning}  & Index \cite{matting_index}  & Copy-paste & Single-Enc & Our\\ \hline\hline
		Average rank  & 5.02 & 4.81  &{4.50} & {3.17} & {3.69} &{4.22} &{4.15}& \textbf{3.05} \\ \hline
	\end{tabular}
	\vskip-6pt

	\label{tab:user}
\end{table*}

 \begin{figure}[t]
	\centering
	\begin{minipage}{.155\textwidth}
		\centering
		\vskip-10pt
		\includegraphics[width=2.6cm,height=3.2cm]{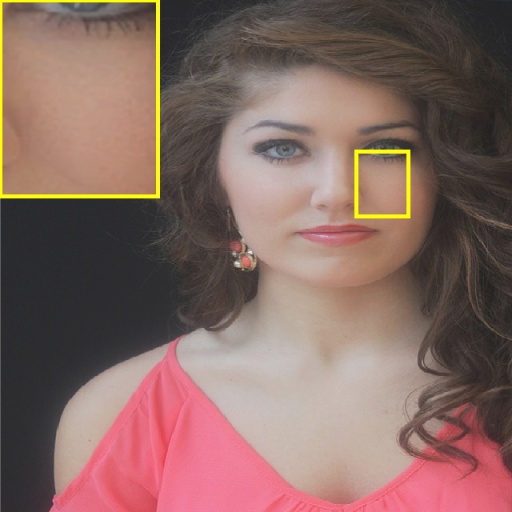}
		\captionsetup{labelformat=empty}
		\captionsetup{justification=centering}
			\vskip-8pt
		\caption*{\emph{Foreground\\ \quad}}
	\end{minipage}
	\begin{minipage}{.155\textwidth}
		\centering
		\vskip-10pt
		\includegraphics[width=2.6cm,height=3.2cm]{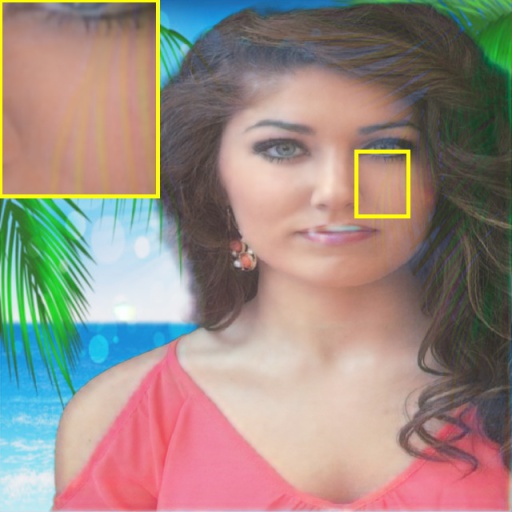}
		\captionsetup{labelformat=empty}
		\captionsetup{justification=centering}
			\vskip-8pt
		\caption*{\emph{w/o-DataAug\\ \quad}}
	\end{minipage}
	\begin{minipage}{.155\textwidth}
		\centering
		\vskip-10pt
		\includegraphics[width=2.6cm,height=3.2cm]{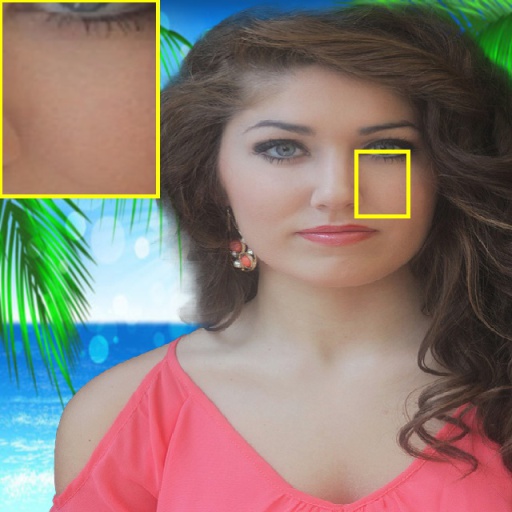}
		\captionsetup{labelformat=empty}
		\captionsetup{justification=centering}
		\vskip-8pt
		\caption*{\emph{Our\\ \quad}}
	\end{minipage}
	\vskip-15pt
	\caption{Results on the  Ablation 1  on a real-world image. Without data augmentation, the baseline MLF network often makes the foreground region mixed with the background colors, leading to color shift artifact.} \label{fig:ablation1}
	\vskip-10pt

\end{figure}

\subsection{Results}

As our task is focusing more on perceptual quality, our evaluation is composed of subject judgment and quantitative evaluation. The quantitative evaluation uses two standard metrics PSNR to demonstrate the compositing quality, and it serves as a verification process. We do not deliberately pursue high scores of these metrics. 

Some visual comparisons are shown in Fig.~\ref{fig:real}\footnote{Due to the page limit, feathering results are put into the supplementary material. }. It can be seen that the Feathering and Laplacian pyramid methods are able to smooth the hard mask and improve the visual quality along the boundary.  However, these two blending methods bring in halo artifacts and the object boundaries tend to be over-smoothed in the composited images. In contrast, the matting-based methods with color decontamination are able to generate compositing results with fine details along the object boundary, but sometimes the boundary artifacts becomes more obvious when the matting is not successful on challenging scenarios, as shown in the third image in Fig.~\ref{fig:real}. In addition, the imperfect trimap generated using the refined segmentation mask may also introduce more difficulty for the matting-based methods, and we find some of matting methods are sensitive to the choice of trimaps. Overall, our method is much more robust to the mistakes in the segmentation mask, and is able to generate higher quality compositing results in most of the cases. It can preserve natural details along the hair and object boundaries, and is also  able to complete the missing part that is not completely captured by the input segmentation mask. Quantitative results evaluated on the synthetic data also demonstrate the effectiveness of the proposed methods, as shown in Table~\ref{tab:state}. The quantitative results in Table~\ref{tab:state} are computed on the unknown regions only. In this setting, our method outperforms Deep Image Matting (DIM) by nearly 1 dB in PSNR and state-of-the-art Index-net \cite{matting_index} by 0.3 dB. Quantitative results evaluated on the whole image region is shown in supplementary material.  Still, the numbers can only partially convey the performance gain of our method. 

\textbf{User Study:}
To further evaluate the perceptual quality of our results, we perform a user study. In this user study, we compare our results with the Laplacian Pyramid Blending method  \cite{blend_laplacian}, and matting based methods using Closed-form  \cite{matting_closed_form}, Information-flow \cite{matting_infor} DIM \cite{matting_ning} and Index \cite{matting_index}. We also include a Single-Stream network (Single-Enc) baseline to verify the architecture design of the MLF network (See the Ablation 2 in Sec~\ref{ssec:ab} for more details). The Copy-paste baseline with refined soft mask is also included to demonstrate the advantage of our deep image composting framework. 

This study involves 44 participants including Photoshop experts. During this study, each participant was shown 14 image sets, each consisting of the foreground images and compositing results of all compared  methods. All testing images and compositing results are included in the supplementary materials. In each image set, participants were asked to pick and rank the favorite 3 results. The composting results in each set is randomly shuffled to avoid selection bias. 
We report the average ranking (lower the better) in Table~\ref{tab:user}. Compositing results that are not selected are assigned a rank score of 8 for more penalty. 

Our method achieves the best ranking score. Among the 14 test samples, our method ranks the first on 9 images. The runner-up, which is Index-net \cite{matting_index}, ranks the first on 4 images. And DIM \cite{matting_ning} is the third place, which ranks the first on 1 image. Other matting-based methods are generally ranked lower than the our baselines. 
One major reason is that they often produce color artifacts along the object boundary, especially on challenging images where color contrast between object and background is not strong enough, or there are strong textures on the background near the object boundary. Moreover, some matting methods are sensitive to the trimaps and their performance may degrade significantly when the trimap is not accurate enough. 
In such cases, the users even prefer the smoother Lap-Pyramid results over the sharper matting-based results. These findings suggest the necessity of an end-to-end formulation for the image compositing problem. 

\begin{figure}[t]
	\centering
	\begin{minipage}{.155\textwidth}
		\centering
	\vskip-8pt
		\includegraphics[width=2.6cm,height=3.2cm]{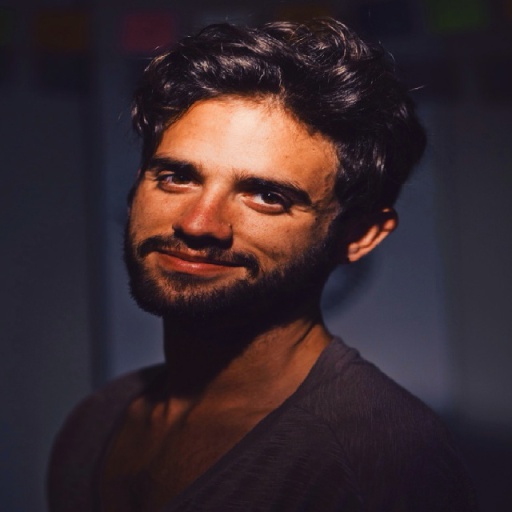}
		\captionsetup{labelformat=empty}
		\captionsetup{justification=centering}
		\vskip-8pt
		\caption*{\emph{Foreground\\ \quad}}
	\end{minipage}
	\begin{minipage}{.155\textwidth}
		\centering
	\vskip-8pt
		\includegraphics[width=2.6cm,height=3.2cm]{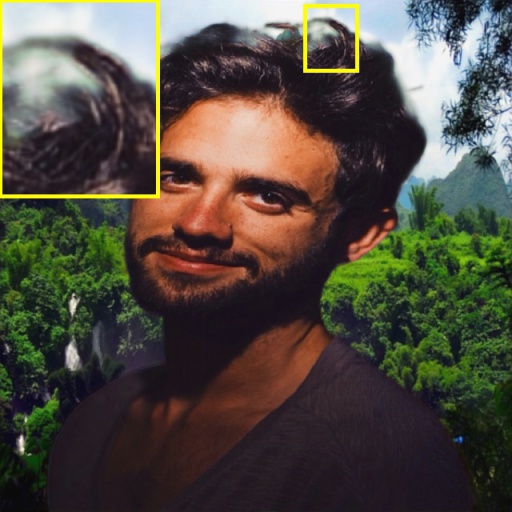}
		\captionsetup{labelformat=empty}
		\captionsetup{justification=centering}
		\vskip-8pt
		\caption*{\emph{Single-Enc\\ \quad}}
	\end{minipage}
	\begin{minipage}{.155\textwidth}
		\centering
	\vskip-8pt
		\includegraphics[width=2.6cm,height=3.2cm]{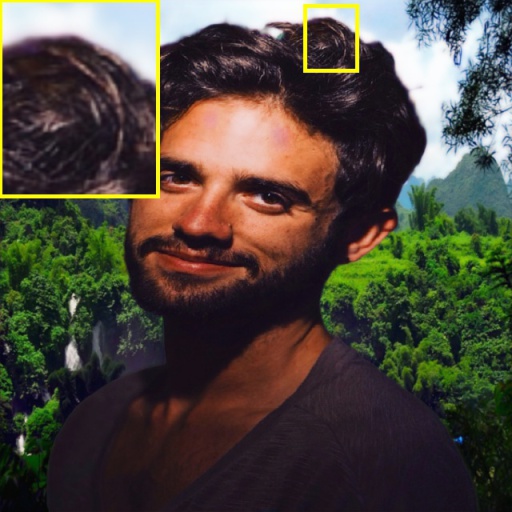}
		\captionsetup{labelformat=empty}
		\captionsetup{justification=centering}
		\vskip-8pt
		\caption*{\emph{Our\\ \quad}}
	\end{minipage}
	\vskip-15pt
	\caption{Results of the  Ablation 2  on a real-world  image. The baseline using a single-stream encoder tends to have issues in preserving the foreground regions.} \label{fig:ablation2}
	\vskip-10pt

\end{figure}


\begin{table}
	\centering
	\footnotesize
	\caption{Quantitative results on \textbf{SynTest} of \emph{Ablation 1-3}, where w/o-DataAug denotes the network trained without our data augmention, Single-Enc denotes a network with a single-stream encoder and w/o-RefNet means the baseline without the segmentation refinement network. Evaluation on unknown regions only.}
			\vskip-6pt
		\begin{tabular}{|c|ccc|c|}
			\hline
			 & w/o-DataAug & Single-Enc & w/o-RefNet & Our \\ \hline\hline
			PSNR (dB)  & {18.00} & {19.05}& {18.22} & \textbf{19.34} \\ \hline
	\end{tabular}
	\label{tab:baselinetable1}
\end{table}





\subsection{Ablation Study} \label{ssec:ab}
We conduct three ablation studies to demonstrate the effectiveness of the proposed method. The quantitative evaluation is performed on the synthesized \textbf{SynTest} datasets. 

\noindent \textbf{Ablation 1: Effectiveness of Data Augmentation}. We evaluate the effectiveness of our self-teach data augmentation method. We train a baseline MLF network only on the matting-based compositing data (see Sec.~\ref{sec:exper}) without using the self-taught based data augmentation (denoted as w/so-DataAug).

By visually checking testing samples, we find that the baseline MLF network is much less robust without the self-taught data augmentation. In many cases, the foreground tend be transparent, leading to color shift on the foreground. A sample result is shown in Fig.~\ref{fig:ablation1}. Quantitative results in Table \ref{tab:baselinetable1} also verify this observation. It shows that large training data is essential for training a robust MLF network and our self-taught data augmentation can effectively mitigate the lack of training data.

\noindent \textbf{Ablation 2: Effectiveness of the Two-stream Encoder}. Next, we demonstrate the effectiveness of our two-stream encoder design. We train a baseline network with a single-stream encoder-decoder structure (denoted as Single-Enc), where the foreground and background image, together with the refined mask, are concatenated as the input to the network. The backbone model of this baseline is the same as our full model.  We do make sure the parameters for both single and two-stream have approximately the
same number of parameters. We increase the number of
channels for the encoder in the single-stream network 

Similar to the Ablation 1, we evaluate this baseline on both the synthetic datasets and real-world images. A visual comparison is shown in Fig.~\ref{fig:ablation2}. It can be observed from the zoom-in region that the single-stream network is less successful in preserving the foreground regions and causes more artifacts along the boundary.  Quantitative results in Table \ref{tab:baselinetable1} is also consistent this observation.


\begin{figure}[t]
	\centering
	\begin{minipage}{.155\textwidth}
		\centering
			\vskip-8pt

		\includegraphics[width=2.6cm,height=3.2cm]{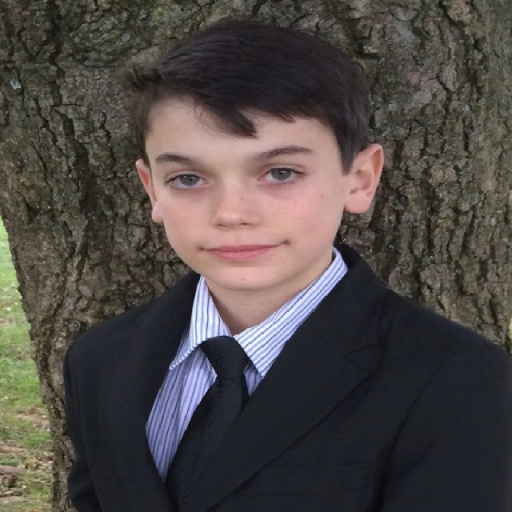}
		\captionsetup{labelformat=empty}
		\captionsetup{justification=centering}
			\vskip-8pt

		\caption*{\emph{Foreground}}
	\end{minipage}
	\begin{minipage}{.155\textwidth}
		\centering
	\vskip-8pt
		\includegraphics[width=2.6cm,height=3.2cm]{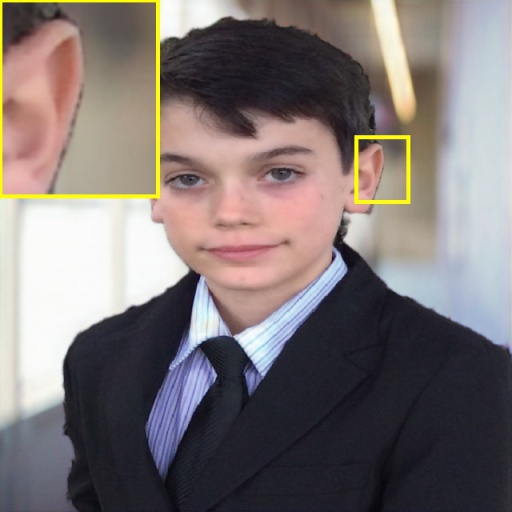}
		\captionsetup{labelformat=empty}
		\captionsetup{justification=centering}
			\vskip-8pt

		\caption*{\emph{w/o-RefNet}}

	\end{minipage}
	\begin{minipage}{.155\textwidth}
		\centering
	\vskip-8pt
		\includegraphics[width=2.6cm,height=3.2cm]{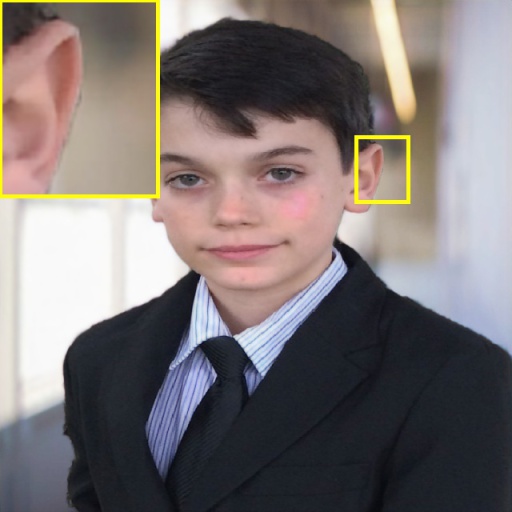}
		\captionsetup{labelformat=empty}
		\captionsetup{justification=centering}
			\vskip-8pt
		\caption*{\emph{Our}}
	\end{minipage}
	\vskip-8pt

	\caption{Results of the  Ablation 3  on a real-world  image. It can be observed from the zoom-in region that the refinement networks enables cleaner boundaries in the composited image.  } \label{fig:ablation3}
	\vskip-10pt

\end{figure}

\noindent \textbf{Ablation 3: Effectiveness of the Mask Refinement Network}. We further evaluate the benefit of introducing the mask refinement network. For this baseline (denoted as w/o-RefNet), we remove the refinement network and directly use the raw segmentation mask for testing. To make it fair, we also re-train the MLF network using the raw mask. 

Sample visual result is shown in Fig.~\ref{fig:ablation3}. In general, the baseline is able to generate high-quality compositing results, but the object boundaries often contain contaminated colors that belong to the original background of the portrait photo. Quantitative results in Table \ref{tab:baselinetable1} echos this observation.

\textbf{Failure Case:}
The proposed MLF compoting network is robust to small errors in the segmentation mask, but it still relies on the general quality of the masks, as indicated in Ablation 3. Most of our failure cases are caused by failures of the segmentation network. 
Sample failure cases are shown in the supplementary material. 


\section{Conclusion}
In this paper, we propose an end-to-end image compositing framework, where a saliency segmentation model with a refinement module is embedded into the network. To efficiently leverage features of  both foreground and background from different scales, a multi-stream fusion network is proposed to generate the final compositing results. Furthermore, a self-taught data-augmentation algorithm is leveraged to augment the current compositing datasets. Experiments evaluated on both synthetic images and real-world images demonstrate the effectiveness of the proposed method compared to other methods. The user-study also show that the proposed method is able to generate better compositing results with good perceptual quality. 



{\small
\bibliographystyle{ieee}
\bibliography{egbib}
}

\end{document}